\documentclass[sigconf]{acmart}

\PassOptionsToPackage{numbers}{natbib}
\usepackage[most]{tcolorbox}
\usepackage{amsmath}
\usepackage{enumitem}  
\usepackage{tabularx}
\usepackage[numbers]{natbib} 
\usepackage{booktabs} 
\usepackage{multirow}
\usepackage{pifont}   
\usepackage{xcolor}
\usepackage[ruled,vlined]{algorithm2e}
\usepackage[utf8]{inputenc}
\usepackage[table]{xcolor} 
\usepackage{siunitx}    
\usepackage{amsfonts}   
\usepackage{caption}
\usepackage{makecell}

\definecolor{orange-web}{RGB}{255, 165, 0}      
\definecolor{sagegreen}{RGB}{138, 179, 137}     
\definecolor{lemonyellow}{RGB}{255, 247, 0}     
\definecolor{skyblue}{RGB}{135, 206, 235}       
\definecolor{coral}{RGB}{255, 127, 80}          
\definecolor{lavender}{RGB}{230, 230, 250}      
\definecolor{mintgreen}{RGB}{152, 255, 152}     
\definecolor{peach}{RGB}{255, 218, 185}         
\definecolor{steelblue}{RGB}{70, 130, 180}      
\definecolor{rosegold}{RGB}{183, 110, 121}      

\colorlet{boxcolor}{steelblue}  

\newtcolorbox{takeawaybox}[1][]{
  enhanced,
  attach boxed title to top left={xshift=4mm,yshift=-2mm},
  colback=boxcolor!10,
  colframe=boxcolor!60,
  colbacktitle=boxcolor!80,
  coltitle=white,
  fonttitle=\bfseries\small,
  boxed title style={size=small, colframe=boxcolor!80, sharp corners},
  sharp corners,
  boxrule=0.8pt,
  left=4pt, right=4pt, top=4pt, bottom=4pt,
  breakable,
  title={#1}
}

\AtBeginDocument{%
  }

\setcopyright{none}
\copyrightyear{2026}
\acmYear{2026}
\acmDOI{}
\acmConference[RecSys ’26 OARS workshop]{ACM Conference on Recommender Systems}{September 27-October 02, 2026}{Minneapolis, MN, USA}
\acmISBN{}

\begin{document}

\title{Decoding ML Decision: An Agentic Reasoning Framework for Large-Scale Ranking System}

\author{Longfei Yun}
  \authornote{Equal contribution.}
  \affiliation{%
    \institution{Meta}
    \city{Menlo Park}
    \country{United States}}
  \email{loyun@meta.com}

  \author{Yihan Wu}
  \authornotemark[1]
  \affiliation{%
    \institution{Meta}
    \city{Menlo Park}
    \country{United States}}
  \email{yihanwu@meta.com}

  \author{Haoran Liu}
  \authornote{Correspondence.}
  \authornotemark[1]
  \affiliation{%
    \institution{Meta}
    \city{Menlo Park}
    \country{United States}}
  \email{ryan0814@meta.com}

\author{Xiaoxuan Liu}
\affiliation{%
  \institution{Meta}
  \city{Menlo Park}
  \country{United States}}
\email{xiaoxuanliu@meta.com}

\author{Ziyun Xu}
\affiliation{%
  \institution{Meta}
  \city{Menlo Park}
  \country{United States}}
\email{vincentxu@meta.com}

\author{Yi Wang}
\affiliation{%
  \institution{Meta}
  \city{Menlo Park}
  \country{United States}}
\email{yiwang1@meta.com}

\author{Yang Xia}
\affiliation{%
  \institution{Meta}
  \city{Menlo Park}
  \country{United States}}
\email{xiay@meta.com}

\author{Pengfei Wang}
\affiliation{%
  \institution{Meta}
  \city{Menlo Park}
  \country{United States}}
\email{pengfeiw@meta.com}

\author{Mingze Gao}
\affiliation{%
  \institution{Meta}
  \city{Menlo Park}
  \country{United States}}
\email{gaomingze@meta.com}

\author{Yunxiang Wang}
\affiliation{%
  \institution{Meta}
  \city{Menlo Park}
  \country{United States}}
\email{yunxiangw@meta.com}

\author{Changfan Chen}
\affiliation{%
  \institution{Meta}
  \city{Menlo Park}
  \country{United States}}
\email{changfanchen@meta.com}

\author{Fei Liu}
\affiliation{%
  \institution{Meta}
  \city{Menlo Park}
  \country{United States}}
\email{feiliu1@meta.com}

\author{Akos Lada}
\affiliation{%
  \institution{Meta}
  \city{Menlo Park}
  \country{United States}}
\email{akos@meta.com}

\author{Wenjie Fu}
\affiliation{%
  \institution{Meta}
  \city{Menlo Park}
  \country{United States}}
\email{jerryf@meta.com}

\author{Hong Yan}
\affiliation{%
  \institution{Meta}
  \city{Menlo Park}
  \country{United States}}
\email{hong@meta.com}

\author{Junfeng Pan}
\authornotemark[2]
\affiliation{%
    \institution{Meta}
    \city{Menlo Park}
    \country{United States}}
\email{panjunfeng@meta.com}

\begin{abstract}

Modern large-scale ranking systems operate within a landscape of competing objectives, operational constraints, and evolving product requirements. Progress in this domain is increasingly bottlenecked not by modeling techniques alone, but by the \emph{engineering context constraint}: the arduous process of translating ambiguous product intent into executable, verifiable hypotheses that account for feature stability, deployment feasibility, and multi-objective trade-offs. We present GEARS (Generative Engine for Agentic Ranking Systems), an orchestration framework that decomposes ranking optimization into three coordinated stages: (1)~programmatic
candidate generation and filtering, which translates natural-language intent into deterministic operations over tabular policy data; (2)~domain-grounded interpretation via Specialized Agent Skills, which provide on-demand access to ranking-specific knowledge such as feature definitions and stability diagnostics; and (3)~deterministic lifecycle governance, which enforces feature-stability and statistical-significance checks before any policy is promoted for deployment. We evaluate GEARS on a benchmark of 100 structured policy-selection instructions derived from 20 large-scale online experiments (0.5B users per experiment, 10B user-experiment pairs total), where it achieves 0.94 nDCG@1, outperforming prompting baselines and code-execution methods. Ablation analysis reveals that programmatic filtering is the primary driver on structured tasks, while specialized skills provide complementary gains on tasks requiring contextual interpretation. Deployment across nine production surfaces demonstrates consistent metric improvements with an annual LLM cost under \$1{,}500 and a reduction in engineering cycle time from weeks to days.
\end{abstract}



\maketitle


\section{Introduction}

Modern ranking systems within large-scale social media platforms
orchestrate heterogeneous product surfaces, ranging from
discovery-oriented recommendation to community-driven social
interfaces, serving a billion-scale global user base with
multifaceted preferences. Over decades of iterative development,
these systems have evolved into highly intricate architectures where
numerous optimization layers concurrently target diverse and often
conflicting metrics. Consequently, the primary bottleneck to system
advancement has shifted from pure signal estimation to the
\emph{engineering context constraint}: the arduous translation of
product intuition and domain expertise into auditable, executable
hypotheses. Current industry workflows remain tethered to manual
intervention, relying on domain experts to navigate the combinatorial
complexity of multi-objective trade-offs, treatment
interpretability, and the rigorous alignment with evolving business
criteria. This manual dependence creates a scalability barrier,
leaving high-value policies undiscovered within the search space.

Furthermore, traditional optimization approaches, such as uplift
modeling~\citep{metalearners,zhao2017uplift,wei2024multi}, treat
personalization as a static model selection task. While these methods
may identify statistically promising interventions, they frequently
fail to account for operational constraints or feature instability,
resulting in policies that are optimal offline but brittle or
undeployable in production environments. We refer to this disconnect
between statistical optimality and production feasibility as the
\emph{deployment gap}.

To bridge this gap, we introduce GEARS (Generative Engine for
Agentic Ranking Systems), a framework that decomposes the end-to-end
ranking optimization workflow into three coordinated stages,
orchestrated by an LLM-based controller. Rather than attempting to
solve for optimality in a single inference step, GEARS treats the
experimentation ecosystem as an interactive environment where each
stage applies the most appropriate tool for the sub-task at hand:

\begin{enumerate}[nosep,leftmargin=1.5em]
\item \textbf{Programmatic candidate generation.} The system
  translates a natural-language product intent into executable
  search specifications, generates candidate policies via
  tolerance-based Pareto filtering, and applies deterministic
  pre-filtering through code execution. This stage handles
  structured, data-manipulation operations where programmatic
  execution is more reliable than text-based reasoning.

\item \textbf{Domain-grounded interpretation.} Specialized Agent
  Skills provide the controller with on-demand access to
  ranking-specific knowledge---feature definitions, stability
  diagnostics, and trade-off explanations---that is absent from
  both the LLM's pretraining data and the tabular experiment
  records. Skills are loaded lazily to bound context length and
  mitigate context rot in long sessions.

\item \textbf{Deterministic lifecycle governance.} Validation hooks
  enforce production safety by checking feature stability and
  statistical significance before any policy is promoted. Only
  candidates that pass all deterministic checks are included in
  the final recommendation set.
\end{enumerate}

This design reflects a deliberate division of labor: code execution
handles what it does best (filtering, sorting, ranking over tabular
data), domain skills supply contextual knowledge that code alone
cannot provide, and governance hooks ensure deployability. The
LLM-based controller orchestrates these heterogeneous components
rather than attempting to replace them.

Experimental validation on a benchmark of 100 structured
policy-selection instructions derived from 20 billion-user-scale
online experiments demonstrates that GEARS achieves 0.94 nDCG@1,
outperforming prompting baselines and code-execution methods.
Ablation analysis confirms that programmatic filtering is the
primary contributor on structured tasks, while specialized skills
provide complementary gains on tasks requiring contextual
interpretation. Deployment across nine production surfaces delivers
consistent metric improvements, reducing the engineering cycle from
weeks to days at an annual LLM cost under \$1{,}500.

\smallskip\noindent Our contributions are threefold:
\begin{enumerate}[nosep,leftmargin=1.5em]
\item \textbf{Orchestration framework for ranking optimization.}
  We decompose ranking policy selection into programmatic
  filtering, domain-grounded interpretation, and deterministic
  governance, coordinated by an LLM-based controller.

\item \textbf{Specialized Agent Skills.} We introduce modular,
  on-demand knowledge resources that externalize ranking expertise
  into reusable, auditable capabilities.

\item \textbf{Production validation at scale.} We demonstrate the
  effectiveness of GEARS across nine production surfaces serving
  billions of users, with quantitative analysis of each
  component's contribution.
\end{enumerate}

\section{Related Works}

\paragraph{The Evolution of Context Engineering}  
LLMs have evolved from simple instruction-following systems into core reasoning engines, necessitating a shift from \textit{prompt engineering} to the formal discipline of \textit{Context Engineering}~\citep{mei2025survey, amatriain2024prompt, ye2024prompt, velsquezhenao2023prompt}. This paradigm shift reconceptualizes the input context not as a monolithic, static string, but as a dynamically structured assembly of informational components.  The development of Tool-Integrated Reasoning (TIR) has further transformed LLMs from passive text generators into \textit{world interactors} capable of autonomous environmental manipulation through structured function calling~\citep{qian2025toolrl, mialon2023augmented, dong2025tool, chen2022program, li2023chain}. Furthermore, contextual self-refinement mechanisms, such as Self-Refine~\citep{madaan2023self} and Reflexion~\citep{shinn2023reflexion}, demonstrate that models can improve their own output quality through iterative feedback loops and reflective episodic memory. 
To address the inherent statelessness of LLMs, research into Memory Systems has introduced OS-inspired hierarchical storage and cognitive architectures. Implementations such as MemGPT~\citep{packer2023memgpt} leverage virtual memory paging to traverse limited context windows, while frameworks like MemoryBank~\citep{zhong2023memory} utilize cognitive principles, such as the Ebbinghaus forgetting curve, to dynamically update memory strength. Within the domain of Multi-Agent Systems (MAS), current research prioritizes sophisticated communication protocols and orchestration mechanisms~\citep{anthropic2024modelcontext, anthropic2025skills, anthropic2025hooks}.

\paragraph{Uplift Modeling and HTE}  Traditional uplift modeling approaches, which seek to identify user segments most responsive to specific interventions, can be broadly categorized into meta-learner based methods, tree-based methods, and neural network-based methods. 
Meta-learner based methods utilize existing prediction models to estimate individual treatment effects (ITE). 
Approaches such as S-learner and T-learner~\citep{metalearners} either combine treatment variables with user features in a single model or build separate models for control and treatment groups, respectively. While effective, these paradigms can suffer from performance degradation when there is a significant imbalance in data between groups. 
Tree-based~\citep{zhao2017uplift,radcliffe2011real,athey2016recursive,nandy2023generalized} methods employ decision trees or forests to partition the user population into subgroups based on their sensitivity to different treatments, using treatment information as part of the splitting criteria. These methods are valued for their interpretability and ability to uncover heterogeneous treatment effects.
Neural network-based methods~\citep{wei2024multi,cevae,bica2020estimating,sun2024m,dragonnet} leverage the representational power of deep learning to model complex user responses and estimate uplift. By introducing flexible architectures, these methods can capture intricate relationships between user features and interventions, and have been widely adopted in domains such as online marketing and precision targeting. Meta has also been investing in this area over a decade, such as Smart Scorer~\cite{peysakhovich2016combining,lada2019observational},but those methods are only intended to solve algorithmic problems, without incorporating any ranking context. 

Despite their strengths, most existing approaches rely heavily on offline features and static user profiles, which limits their ability to respond to real-time shifts in user behavior or to dynamically adapt intervention strategies. In practice, even when an algorithm delivers the “best” offline result, it may never ship, because large-scale ranking systems are highly complex. Any deployable solution must account for business objectives, existing ranking policies, and cannibalization across products and systems. Furthermore, the design and deployment of these models often require manual workflows that are time-consuming and resource-intensive.

\paragraph{Adaptive Experimentation (AE)} Distinct from static prediction models, Adaptive Experimentation (AE) focuses on optimizing outcomes through sequential decision-making and active exploration. AX methodologies, including Multi-Armed Bandits and Bayesian Optimization, are designed to balance exploration and exploitation to identify optimal configurations or policies dynamically. In the context of large-scale ranking, AE has been utilized to automate the tuning of system parameters and personalization strategies. For instance, recent works at Meta~\citep{olson2025ax,wu2022interpretable,bakshy2018ae} demonstrate how AE can be deployed to efficiently search vast parameter spaces, allowing ranking systems to adapt to user feedback more rapidly than traditional A/B testing cycles would permit.

\section{Preliminary}
\subsection{Notation}
\paragraph{Experimental Setup and Notation}
We consider a randomized online experiment (A/B test) with a set of $M$ actions (treatments)
\[
\mathcal{A}=\{a_1,\ldots,a_M\},
\]
and a control action $a_0$.
We evaluate $K$ performance metrics $\{\delta_1,\ldots,\delta_K\}$.
Each user $u_i$ is associated with a $D$-dimensional feature vector
\[
\{X_1(u_i),\ldots,X_D(u_i)\},\quad X_d(u_i)\in\mathcal{X}.
\]
Users are randomly assigned to one of the treatment groups $\{U_1,\ldots,U_M\}$ or the control group $U_0$.
We write $\delta_k(u_i,a_j)$ for the observed outcome of user $u_i$ under action $a_j$ on metric $\delta_k$.

\paragraph{Average Treatment Effect (ATE)}
The average treatment effect of treatment $a_j$ relative to control $a_0$ on metric $\delta_k$ is
\small{
\[
\mathrm{ATE}(a_j,\delta_k)
=
\frac{1}{|U_j|}\sum_{u_i\in U_j}\delta_k(u_i,a_j)
-
\frac{1}{|U_0|}\sum_{u_i\in U_0}\delta_k(u_i,a_0).
\]
}

\paragraph{Heterogeneous Treatment Effect (HTE)}
The individual-level heterogeneous treatment effect of treatment $a_j$ relative to control $a_0$ for user $u_i$ on metric $\delta_k$ is
\[
\mathrm{HTE}(u_i,a_j,\delta_k)
=
\delta_k(u_i,a_j)-\delta_k(u_i,a_0).
\]
By leveraging flexible architectures, HTE are able to model complex interactions between user features and interventions, and have seen widespread adoption in areas like online marketing and precision targeting.

\subsection{GAS}
\label{sec:gas}
GAS~(\underline{G}eneralized \underline{A}utomatic
\underline{S}egmentation)~\citep{wu2025gas} is a user-segment level HTE algorithm to improve personalization strategies and overcome obstacles. GEARS builds its candidate generation stage on top of GAS.

Let $\mathcal{S}$ denote a collection of non-overlapping user segments, and let $S\in\mathcal{S}$ be a segment (a subset of users).
The segment-level heterogeneous treatment effect of $a_j$ relative to $a_0$ on metric $\delta_k$ is
\[
\mathrm{HTE}(S,a_j,\delta_k)
=
\frac{1}{|S|}\sum_{u_i\in S}\left(\delta_k(u_i,a_j)-\delta_k(u_i,a_0)\right).
\]

\paragraph{Policy Parameterization through Quantile-Based Segmentation.}
GAS defines candidate policies through quantile-based cohort segmentation over user features. Let $Q(X,p)$ denote the $100p$-th percentile of feature $X\in\mathcal{X}$, with $Q(X,0)=-\infty$. Let $N\in\mathbb{Z}_{+}$ denote the number of quantile bins.

\emph{Individual split.}
For feature $X$, define $N$ segments:
\small{
\[
S^{\mathrm{ind}}_{i}(X)
=
\left\{
u \,\middle|\,
Q\!\left(X,\frac{i-1}{N}\right) < X(u) \le Q\!\left(X,\frac{i}{N}\right)
\right\}.
\]
}

\emph{Binary split.}
For threshold index $i_0\in\{1,\ldots,N-1\}$:
\[
S^{\mathrm{bin}}_{1,i_0}(X)
=
\left\{
u \,\middle|\,
Q(X,0) < X(u) \le Q\!\left(X,\frac{i_0}{N}\right)
\right\},
\]
\[
S^{\mathrm{bin}}_{2,i_0}(X)
=
\left\{
u \,\middle|\,
Q\!\left(X,\frac{i_0}{N}\right) < X(u) \le Q(X,1)
\right\}.
\]

These segmentation strategies define a large combinatorial space of candidate policies across cohorts and treatments.

\paragraph{Finding Pareto Policies}
A policy is a mapping from segments to actions and can be represented as a set of $(\text{segment},\text{action})$ pairs:
\[
\mathcal{P}=\{(S_1,a_1),\ldots,(S_B,a_B)\},
\]
where the segments $\{S_b\}_{b=1}^B$ form a partition of the user population.

In order to identify Pareto-optimal policies that jointly optimize multiple metrics, GAS adopts an efficient \emph{weight-search} procedure. Specifically, it defines a family of \emph{linear scalarized rewards} over $K$ metrics using a set of weight vectors $\mathbf{W} \subset \mathbb{R}^K$. For each segment $S_b$ (with associated action/policy $a_b$), GAS evaluates the weighted average metric lift $\delta_k(u_i, a_b)$ across users $u_i \in S_b$, and selects the set of policies induced by maximizing this scalarized objective over $\mathbf{w} \in \mathbf{W}$. The resulting Pareto policy set is
\[
\mathcal{P}_{\mathrm{pareto}}
:=
\left\{
\arg\max_{a}
\sum_{b=1}^{B}\sum_{k=1}^{K}\frac{w_k}{|S_b|}\sum_{u_i \in S_b} \delta_k(u_i, a_b)
\;\middle|\;
\mathbf{w} \in \mathbf{W}
\right\}.
\]

\begin{figure}[htbp] 
    \centering
     \includegraphics[width=0.49\textwidth]{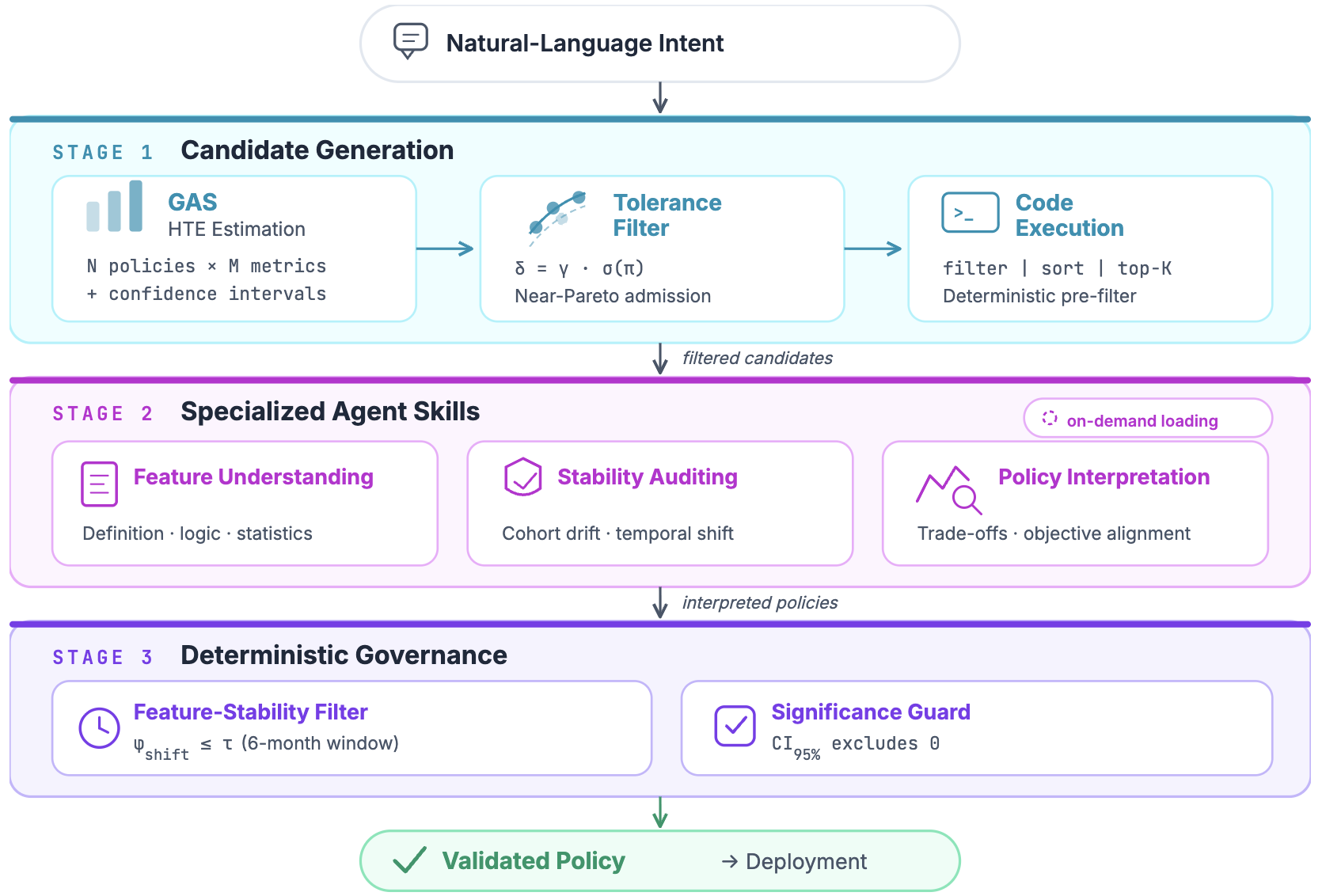}
     \caption{End-to-end GEARS pipeline. The system accepts a natural-language intent, generates and filters candidate policies through programmatic execution (Stage 1), interprets them using domain-specific skills loaded on demand (Stage 2), and validates production readiness through deterministic governance hooks (Stage 3).}
    \label{fig:gears2}
\end{figure}

\section{The GEARS System}

\paragraph{Motivation.}
\begin{figure}[htbp] 
\centering
 \includegraphics[width=0.49\textwidth]{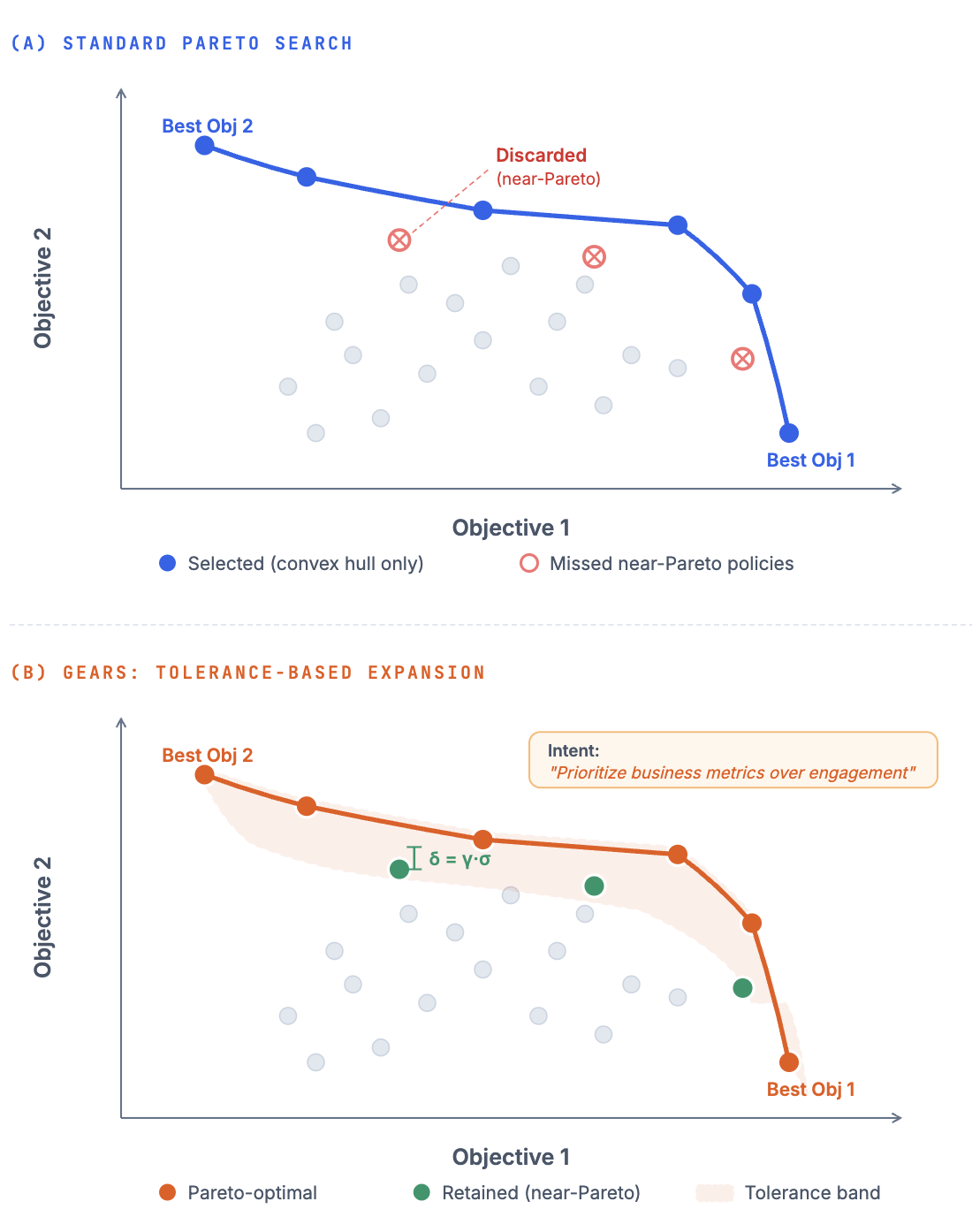}
 \caption{Tolerance-based frontier expansion. (a) Standard Pareto search selects only convex-hull solutions, discarding near-optimal candidates in non-convex regions. (b) GEARS admits policies within an uncertainty-aware tolerance band $\delta = \gamma \cdot \sigma$, retaining near-Pareto candidates that may offer better stability or deployability.}
\label{fig:tolerance}
\end{figure}

Existing personalization frameworks, including deep uplift
models~\citep{metalearners,wei2024multi} and generative
recommendation approaches, treat ranking optimization as static signal estimation. These methods can identify statistically promising interventions but operate without access to \emph{engineering context}:
infrastructure constraints, feature-stability properties, and deployment criteria that determine whether a policy is feasible in production.
For instance, an HTE model may surface a high-lift cohort defined by a feature that exhibits high temporal drift, yielding a policy that is optimal offline but undeployable. 
We refer to this disconnect between statistical optimality and production feasibility as the \emph{deployment gap}.

GEARS bridges this gap by decomposing the end-to-end optimization workflow into three coordinated stages: programmatic filtering over the candidate space, domain-knowledge retrieval through specialized agent skills, and deterministic validation via lifecycle governance hooks. As illustrated in~\autoref{fig:gears2}, an LLM-based controller orchestrates these stages, accepting a natural-language product intent as input and producing a validated, deployment-ready policy configuration.

\paragraph{Intent-Conditioned Candidate Generation.}
The first stage translates a natural-language objective (e.g.,
\emph{``maximize retention without degrading latency''}) into an
executable search over the policy space.

\begin{itemize}
\item \smallskip\noindent\textbf{Tolerance-Based Frontier Expansion.}
GEARS builds on GAS (\S ~\ref{sec:gas}), which estimates
heterogeneous treatment effects across quantile-based user segments and enumerates candidate policies via random-weight Pareto search. Standard random-weight search converges on the convex hull of the Pareto frontier, potentially discarding non-convex candidates that may be preferable for deployment. GEARS introduces a tolerance-based Pareto filter (\autoref{fig:tolerance}; pseudocode in
Algorithm~\ref{alg:tgas}): for each metric~$m$ and candidate policy~$\pi$, a tolerance margin
$\delta_m(\pi) = \gamma \cdot \sigma_m(\pi)$ is defined, where $\sigma_m(\pi)$ is the estimated standard error and $\gamma \geq 0$ is a hyperparameter. A candidate is retained unless another candidate dominates it on every metric even after accounting for these margins. This expands the candidate set to include near-Pareto policies whose stability or interpretability may favor deployment.

\begin{algorithm}[htbp]
\caption{Tolerance-based Personalization: Random-weight Top-$K$ + Tolerance Pareto Filtering}
\label{alg:tgas}
\KwIn{
Policy set $\mathcal{P}$; metrics $m=1,\dots,M$ (assume maximize); estimated means $\mu_m(p)$ and uncertainties $\sigma_m(p)$ for each $p\in\mathcal{P}$;
number of random weights $W$; top-$K$ per weight $K$; tolerance hyperparameter $\tau \ge 0$.
}
\KwOut{Final candidate policy set $\mathcal{C}_\tau$ (Pareto-optimal and near-Pareto policies).}
\vspace{2mm}
\textbf{Step 1: Candidate collection via random weight search.}\\
Sample $W$ weight vectors $\{w^{(i)}\}_{i=1}^W$ from the simplex $\Delta^{M-1}$\;
Initialize candidate set $\mathcal{C} \leftarrow \emptyset$\;
\For{$i \leftarrow 1$ \KwTo $W$}{
    \ForEach{$p \in \mathcal{P}$}{
        $S_{w^{(i)}}(p) \leftarrow \sum_{m=1}^M w^{(i)}_m \, \mu_m(p)$\;
    }
    Let $\mathcal{T}^{(i)} \leftarrow \textsc{TopK}\big(\{S_{w^{(i)}}(p)\}_{p\in\mathcal{P}}, K\big)$ \tcp*[r]{Top-$K$ policies by weighted score}
    $\mathcal{C} \leftarrow \mathcal{C} \cup \mathcal{T}^{(i)}$\;
}
\vspace{2mm}
\textbf{Step 2: Tolerance-based Pareto filtering (near-frontier admission).}\\
Define per-metric tolerance margin for candidate $p$:
\[
\epsilon_m(p) \triangleq \tau \cdot \sigma_m(p).
\]
Define \emph{tolerance-dominance} $q \succ_{\tau} p$ (maximize case) as:
\[
\Big(\forall m,\; \mu_m(q) \ge \mu_m(p) - \epsilon_m(p)\Big)
\;\;\wedge\;\; \]

\[\Big(\exists m,\; \mu_m(q) > \mu_m(p) + \epsilon_m(p)\Big).
\]
Initialize $\mathcal{C}_\tau \leftarrow \emptyset$\;
\ForEach{$p \in \mathcal{C}$}{
    dominated $\leftarrow \textbf{false}$\;
    \ForEach{$q \in \mathcal{C}$}{
        \If{$q \neq p$ \textbf{and} $q \succ_{\tau} p$}{
            dominated $\leftarrow \textbf{true}$\;
            \textbf{break}\;
        }
    }
    \If{\textbf{not} dominated}{
        $\mathcal{C}_\tau \leftarrow \mathcal{C}_\tau \cup \{p\}$\;
    }
}
\Return $\mathcal{C}_\tau$\;
\end{algorithm}

\item \smallskip\noindent\textbf{Programmatic Pre-Filtering.}
Before LLM-based reasoning, the agent translates the user's intent into deterministic operations over the candidate table by generating and executing shell commands. For example, given
\emph{``maximize Metric\,1 without regressing Metric\,2,''} the agent produces a command that retains rows where the lower confidence bound of Metric\,2 is non-negative, then sorts by Metric\,1 in descending order.
This step is deliberately non-neural: structured filtering over tabular data is more reliably handled by programmatic execution than by text-based reasoning, and applying it early reduces the candidate set that the subsequent reasoning stage must process.
As the ablation study in~\S\ref{sec:gears_pass_rate} confirms, this programmatic stage contributes substantially to overall performance on structured selection tasks, consistent with prior findings that code-based execution outperforms pure prompting for data-manipulation
operations~\citep{wang2024executable}.
\end{itemize}
\paragraph{Domain-Grounded Interpretation via Specialized Agent Skills.}
While programmatic filtering handles well-defined, structured queries effectively, many practical decisions require contextual judgment that cannot be reduced to deterministic rules. For instance, determining whether a feature is suitable for long-term targeting, or explaining why a particular cohort split yields a favorable trade-off, requires domain-specific knowledge absent from both the LLM's pretraining data
and the tabular experiment records.

GEARS encapsulates this knowledge in \emph{Specialized Agent Skills}:
modular, read-only resources that provide the agent with structured
procedures and access to internal tools. Each skill consists of
(i)~a \emph{metadata header} describing its purpose and trigger
conditions, and (ii)~a \emph{core body} containing domain logic and
tool-invocation templates. At inference time, only skill metadata is
loaded into the agent's context; the core body is loaded on demand when
the agent determines relevance, bounding context length and mitigating
context rot in long sessions.

Three categories of skills are deployed:
\begin{itemize}[nosep,leftmargin=1.2em]
\item \textbf{Feature understanding} -- retrieves the definition,
computation logic, and historical statistics of a ranking feature,
enabling assessment of whether a policy's reliance on a given feature is operationally sound.
\item \textbf{Stability auditing} -- queries internal monitoring tools to compute the user-cohort shift ratio (\S\ref{sec:shift_ratio}) for each feature in a candidate policy, flagging features with high temporal or distributional drift.
\item \textbf{Policy interpretation} -- summarizes the trade-off profile of a recommended policy and generates a natural-language rationale for human review before deployment.
\end{itemize}

\noindent The contribution of skills is complementary to programmatic
filtering: they improve performance on tasks that require contextual
interpretation beyond structured data manipulation.

\paragraph{Deterministic Lifecycle Governance.}
The final stage ensures that policies surfaced by the preceding
components are robust enough for production deployment. GEARS registers
\emph{validation hooks}---deterministic checks executed before a
candidate is promoted to the recommendation set.

Two hooks are enforced by default:
\begin{itemize}[nosep,leftmargin=1.2em]
\item \textbf{Feature-stability filter.} For every feature in a
candidate policy, the hook computes the \emph{user-cohort shift
ratio}~$\phi_{\mathrm{shift}}$: the fraction of users who migrate
across cohort boundaries over a six-month window. A policy is
admitted only if all its features satisfy
$\phi_{\mathrm{shift}} \leq \tau$, where $\tau$ is calibrated
against an empirically stable baseline feature set
\item \textbf{Statistical-significance guard.} The hook verifies
that reported metric lifts are statistically significant (95\%
confidence interval excludes zero for the primary metric),
filtering candidates whose gains may be attributable to sampling
noise.
\end{itemize}

\noindent Because all hooks are deterministic and logged, they produce
an auditable record of admission and rejection decisions, supporting
both automated governance and human oversight.

\subsection{Illustrative Agent Session}
\begin{figure}[htbp] 
    \centering
     \includegraphics[width=0.49\textwidth]{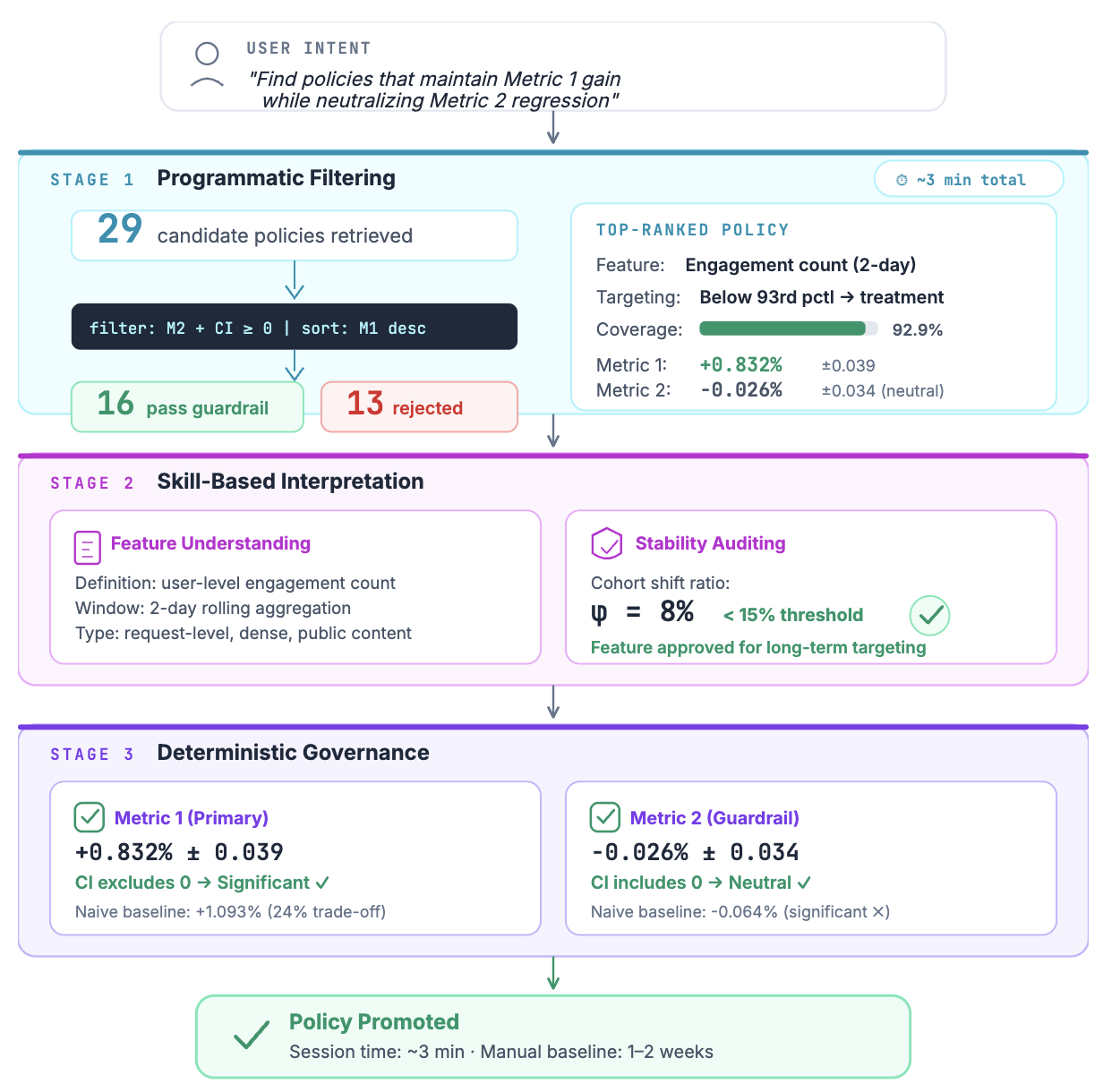}
     \caption{Anonymized trace of a production GEARS session. The system filters 29 candidates to 16 via programmatic guardrail checks (Stage 1), verifies feature stability through domain-specific skills (Stage 2), and validates statistical significance before promoting the policy (Stage 3). The targeted policy neutralizes the guardrail regression present in the naive baseline.}
    \label{fig:session}
\end{figure}

~\autoref{fig:session} presents an anonymized trace of a
production GEARS session. Given the intent \emph{``maintain
Metric~1 gain while neutralizing Metric~2 regression,''} the
pipeline proceeds as follows. In Stage~1, GAS generates hundreds of candidate policies from a billion-user experiment; the tolerance-based Pareto filter retains 29 near-optimal candidates, and programmatic code execution applies the guardrail constraint (Metric~2 value $+$ CI $\geq 0$), reducing the set to 16. In Stage~2, the agent invokes the \emph{feature understanding} skill to retrieve the definition of the top policy's segmentation feature (a two-day engagement count) and the \emph{stability auditing} skill to verify its cohort shift ratio ($\phi_{\mathrm{shift}} = 8\%$, below the $15\%$ threshold). In Stage~3, governance hooks confirm that Metric~1 is statistically significant ($+0.832\% \pm 0.039$) and Metric~2 is neutral ($-0.026\% \pm 0.034$, CI includes zero). The resulting policy retains 76\% of the naive baseline's Metric~1 gain while fully eliminating the guardrail violation. 

\section{Experiments}
In this section, we investigate the following research questions:
\begin{itemize}[nosep,leftmargin=1.2em]
\item \textbf{RQ1}: How does each component of the GEARS pipeline contribute to policy selection performance?
\item \textbf{RQ2}: Does the governance layer improve the
  deployability of recommended policies?
\item \textbf{RQ3}: Can GEARS deliver measurable metric
  improvements in production deployment?
\end{itemize}

\subsection{Experimental Settings}
\label{sec:setup}
\paragraph{Dataset} 
We constructed a benchmark dataset from 20 internal large-scale industry online experiments deployed on a live production system serving billions of users, with approximately 0.5 billion users per experiment and 10 billion user-experiment pairs in total (with overlap across experiments).
For each experiment, we initially ran the GAS~\citep{wu2025gas} algorithm to generate hundreds of policy candidates with their corresponding metric measurements.
To enable objective evaluation with deterministic ground truth, we
defined five instruction templates representing common policy
selection scenarios: 
\begin{itemize} [itemsep=0pt, topsep=2pt, parsep=0pt]      
\item \textbf{Maximize Both}: Find policies that jointly optimize two metrics            
\item \textbf{Maximize with Constraint}: Optimize a primary metric while ensuring a secondary metric does not regress                                        
\item \textbf{Tradeoff Analysis}: Identify Pareto-optimal policies representing different tradeoff points between competing  metrics                                             
\item \textbf{Efficiency Optimization}: Select policies with the highest composite efficiency score                          

\item \textbf{Single Metric}: Maximize a single target metric regardless of others                                           
\end{itemize}                  

Instantiating each template across 20 experiments yields 100 structured instructions. For each, the ground-truth ranking is computed deterministically from the optimization criteria, enabling unambiguous evaluation.

These structured instructions are well suited to programmatic
execution; the purpose of this benchmark is to measure the
reliability of the full pipeline on tasks with verifiable answers,
not to evaluate open-ended reasoning.      

\paragraph{Baselines} 
To assess the effectiveness of our proposed GEARS framework, we benchmark its performance against several established prompting  strategies:                                                
\begin{itemize} [itemsep=0pt, topsep=2pt, parsep=0pt]                                                      
\item \textbf{Naive Prompting}: Directly queries the LLM with the task instruction and data without additional reasoning guidance.                                                            
\item \textbf{Chain-of-Thought (CoT)}~\citep{wei2022chain}: Encourages step-by-step reasoning by prompting the model to first
understand the objective, analyze the data, apply selection criteria, and then provide recommendations.                         
\item \textbf{Self-Consistency}~\citep{wang2022self}: Samples multiple reasoning paths with temperature-based     
diversity ($T=0.7$) and aggregates predictions via Borda count voting to improve robustness.                                     
\item \textbf{Self-Refine}~\citep{madaan2023self}: A two-stage approach where the model first generates initial recommendations, then       
critically reviews and refines its own output to correct potential errors.                            

\item \textbf{Code-as-Action}~\citep{wang2024executable}: Instead of only generating text, the LLM generates and executes code to solve the task. This makes outputs verifiable and reproducible, reducing errors and hallucinations in computation- or data-driven settings. 
\end{itemize}                                           
\begin{table*}[htbp]
\centering
\small
\caption{Quantitative comparison of policy selection performance. We evaluate the proposed GEARS framework against five state-of-the-art baselines. Performance is measured across five key dimensions: Ranking Quality (nDCG@$k$), Precision (Prec@$k$), Global Rank Correlation, Recall (Rec@$k$), and Top-1 Performance.}
\label{tab:policy_selection_csv}
\setlength{\tabcolsep}{6pt}
\renewcommand{\arraystretch}{1.2}

\resizebox{.99\textwidth}{!}{
\begin{tabular}{lcccccccccccc}
\toprule

\textbf{Method}
& \multicolumn{3}{c}{\textbf{Ranking Quality}}
& \multicolumn{3}{c}{\textbf{Precision@k}}
& \multicolumn{1}{c}{\textbf{Global}}
& \multicolumn{3}{c}{\textbf{Recall@k}}
& \multicolumn{2}{c}{\textbf{Top-1 Performance}} \\

\cmidrule(lr){2-4}
\cmidrule(lr){5-7}
\cmidrule(lr){8-8}
\cmidrule(lr){9-11}
\cmidrule(lr){12-13}

& \textbf{nDCG@1}
& \textbf{nDCG@3}
& \textbf{nDCG@5}
& \textbf{Prec@1}
& \textbf{Prec@3}
& \textbf{Prec@5}
& \textbf{Rank Corr.}
& \textbf{Rec@1}
& \textbf{Rec@3}
& \textbf{Rec@5}
& \textbf{Top-1 Acc}
& \textbf{Top-1 in GT} \\

\midrule

Naive & 0.57 & 0.70 & 0.74 & 0.57 & 0.39 & 0.28 & 0.20 & 0.36 & 0.67 & 0.77 & 0.44 & 0.57 \\     

CoT~\cite{wei2022chain} & 0.68 & 0.80 & 0.83 & 0.68 & 0.44 & 0.30 & 0.31 & 0.43 & 0.73 & 0.82 & 0.57 & 0.68 \\  

Self-Consistency~\cite{wang2022self} & 0.57 & 0.74 & 0.76 & 0.57 & 0.39 & 0.28 & 0.13 & 0.33 & 0.67 & 0.78 & 0.37 & 0.57 \\ 

Self-Refine~\cite{madaan2023self}  & 0.61 & 0.78 & 0.80 & 0.61 & 0.44 & 0.31 & 0.34 & 0.37 & 0.74 & 0.83 & 0.50 & 0.61 \\   

Code-as-Action ~\cite{wang2024executable}     
& 0.77
& 0.87
& 0.87
& 0.77
& 0.52
& 0.33
& 0.59
& 0.45
& 0.84
& 0.88
& 0.68
& 0.77 \\

\midrule
GEARS w/o Bash 
& 0.40
& 0.42
& 0.42
& 0.40
& 0.18
& 0.11
& 0.80
& 0.24
& 0.32
& 0.32
& 0.26
& 0.40 \\
  
GEARS w/o Skill 
& 0.87
& 0.91
& 0.91
& 0.87
& 0.57
& 0.35
& 0.72
& 0.53
& 0.89
& 0.90
& 0.77
& 0.87
\\

Ours (GEARS)
& \cellcolor{lavender} \textbf{0.94}
& \cellcolor{lavender} \textbf{0.96}
& \cellcolor{lavender} \textbf{0.96}
& \cellcolor{lavender} \textbf{0.94}
& \cellcolor{lavender} \textbf{0.60}
& \cellcolor{lavender} \textbf{0.37}
& \cellcolor{lavender} \textbf{0.82}
& \cellcolor{lavender} \textbf{0.56}
& \cellcolor{lavender} \textbf{0.94}
& \cellcolor{lavender} \textbf{0.95}
& \cellcolor{lavender} \textbf{0.86}
& \cellcolor{lavender} \textbf{0.94} \\
\bottomrule
\end{tabular}
}
\end{table*}

\paragraph{Metrics}
To evaluate policy selection performance, we employ standard ranking and retrieval metrics widely adopted in information retrieval and recommender systems.

\begin{itemize} [itemsep=0pt, topsep=2pt, parsep=0pt]
\item \textbf{Precision@K} measures the fraction of recommended policies within the top-K that are included in the ground-truth set. 
\item \textbf{Recall@K} quantifies the proportion of ground-truth policies that appear within the top-K predictions.       

\item \textbf{NDCG@K} (Normalized Discounted Cumulative Gain) evaluates not only whether the model retrieves relevant policies but also whether they are ranked near the top: 
\small{
\[\mathrm{NDCG@K} = \frac{\mathrm{DCG@K}}{\mathrm{IDCG@K}}, \mathrm{DCG@K} = \sum_{i=1}^{K} \frac{2^{\mathrm{rel}_i}-1}{\log_2(i+1)},
\]}
where $\mathrm{rel}_i \in \{0, 1\}$ indicates whether the policy at position $i$ belongs to the ground-truth set.            
\item \textbf{Top-1 Accuracy} measures whether the highest-ranked prediction exactly matches the best ground-truth policy.   
\item \textbf{Top-1 in GT} reports whether the top-ranked prediction belongs to the ground-truth set (a relaxed version of   
Top-1 Accuracy).                                        \item \textbf{Ranking Correlation} assesses the agreement between predicted and ground-truth rankings via Spearman's $\rho$, 
capturing global ordering fidelity.                     \end{itemize}      

We report results for $K \in \{1, 3, 5\}$. Together, these metrics capture complementary aspects of performance: correctness     
(Top-1 Accuracy), coverage (Recall@K), precision-coverage tradeoff (Precision@K), ranking quality (NDCG@K), and global ordering  (Ranking Correlation). 

\paragraph{Implementation Details}
All experiments use Claude Sonnet 4.6~\citep{anthropicsonnet} as the backbone LLM. For Self-Consistency, we sample 5 responses per instruction with          
temperature 0.7 and aggregate via Borda count. For Self-Refine, we use a single refinement iteration.     

\subsection{Structured Policy Selection and Component Analysis}
\label{sec:gears_pass_rate}

We evaluate GEARS under an \emph{offline policy selection} setting, where the model is given tabular experiment records of multiple candidate policies and is instructed to output a ranked list of recommended policies. Each candidate policy is associated with multiple metrics (e.g., primary objective and guardrail metrics).

~\autoref{tab:policy_selection_csv} reports the policy selection performance across a suite of ranking and decision-oriented metrics. GEARS consistently outperforms all baselines across most metrics, indicating stronger reliability in selecting high-quality policies under multi-metric constraints.

We further conduct ablation studies to isolate the contribution of each component in GEARS. On structured instructions, removing code execution (GEARS w/o Bash) causes the largest performance drop (nDCG@1: 0.94 $\to$ 0.40), confirming that programmatic filtering is the primary driver for deterministic selection tasks---an expected result, since these instructions map directly to tabular operations (filter, sort, rank) that are more reliably executed as code than reasoned about in text.

Removing skills (GEARS w/o Skill) yields a smaller but consistent drop (nDCG@1: 0.94 $\to$ 0.87). On structured tasks, skills provide incremental gains by improving the agent's interpretation of domain-specific selection criteria. In production, where instructions are less structured and feature context is critical, the contribution of skills is substantially larger.

\subsection{Hooks Improve the Reliability of GEARS Recommendation}
\label{sec:shift_ratio}
\begin{table}[h]
\centering
\caption{Feature Stability Benchmark (6-Month Window). Binary cuts align engagement feature stability with feature set $\mathcal{S}$ baselines.}
\label{tab:stability_benchmarks}
\resizebox{.49\textwidth}{!}{
\begin{tabular}{lcccc}
\toprule
\textbf{Feature Class} & \textbf{Feature Name} & \textbf{Shift (Quantile)} & \textbf{Shift (Binary)} & \textbf{Status} \\
\midrule
Baseline & Feature Set $\mathcal{S}$ & 6\% & 2\% & \textcolor{green!60!black}{Benchmark} \\
Product & Feature 2 & 16\% & 4\% & \textcolor{green!60!black}{Stable} \\
Engagement & Feature 3 & 30\% & 10-12\% & \textcolor{green!60!black}{Stable (Binary)} \\
Product & Feature 4 & $\sim$50\% & $\sim$20\% & \textcolor{red!60!black}{Unstable} \\
Product & Feature 5 & N/A & $\sim$30\% & \textcolor{red!60!black}{Unstable} \\
\bottomrule
\end{tabular}
}
\end{table}
To ensure that the policies generated by GEARS are deployable and robust against temporal distribution shifts, we established a rigorous quantitative benchmark for checking \textit{feature stability}.

\subsubsection{Benchmarking Methodology}
We define the \textit{User-Cohort Shift Ratio} ($R_{\text{shift}}$) as the primary metric for stability: the percentage of users who migrate from their assigned cohort (bucket) to a different one over a 6-month window.
To set a realistic baseline, we benchmarked a \textit{perceived stable} feature set $\mathcal{S}$ using two cohort definitions: Quantile Cuts (4 equal buckets) and Binary Cuts (p25/p75 thresholds).

\begin{figure}[htbp] 
    \centering
     \includegraphics[width=0.49\textwidth]{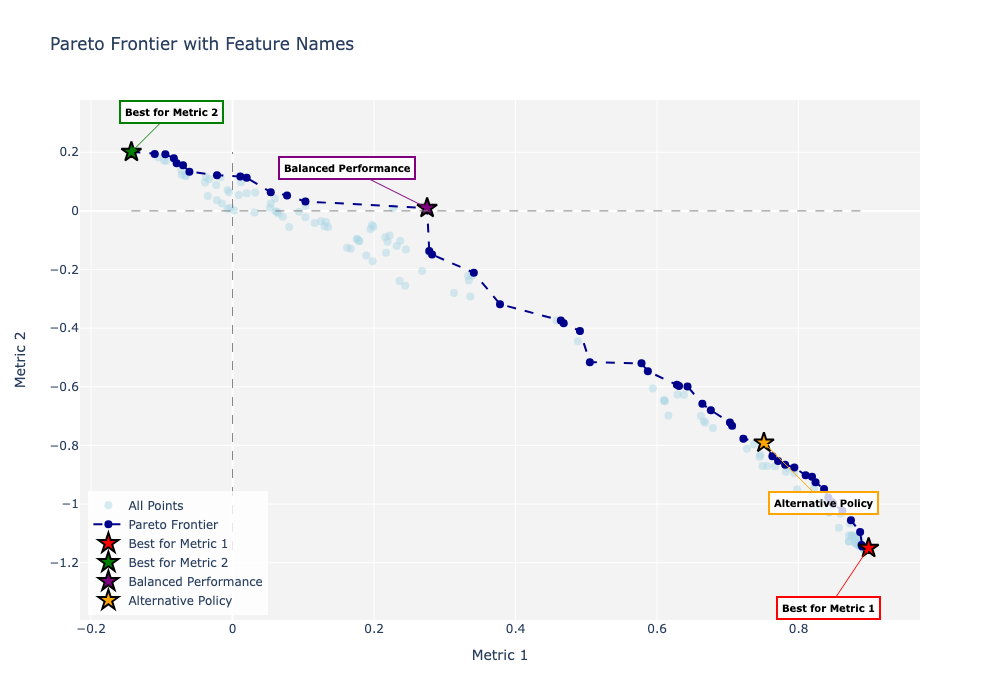}
     \caption{Pareto efficiency of generated policies. We plot the performance of all candidate policies, with the Pareto frontier (dark blue line) indicating the optimal trade-off curve. Annotated stars mark key policies of interest.}
    \label{fig:stability}
\end{figure}

\subsubsection{Empirical Baselines and Thresholds}
Our analysis revealed that even the baseline  exhibits drift. As shown in ~\autoref{tab:stability_benchmarks}, feature set $\mathcal{S}$ shifted by $6\%$ over 6 months, establishing a lower bound for unavoidable natural drift.
Product features, such as \textit{Feature 4}, showed high volatility ($50\%$ shift) under quantile cuts, making them unsuitable for long-term policy targeting.

Based on these benchmarks, we implemented a validation hook within GEARS:
\begin{itemize}
    \item \textbf{Pre-Search Filter:} Features must exhibit $R_{\text{shift}} \le 15\%$ (Binary) or $\le 45\%$ (Quantile) to enter the search space.
\end{itemize}

This benchmarking process allowed GEARS to automatically disqualify high-lift but unstable features (e.g., Feature 4) that baseline methods would have erroneously selected.

\begin{figure}[htbp] 
    \centering
     \includegraphics[width=0.49\textwidth]{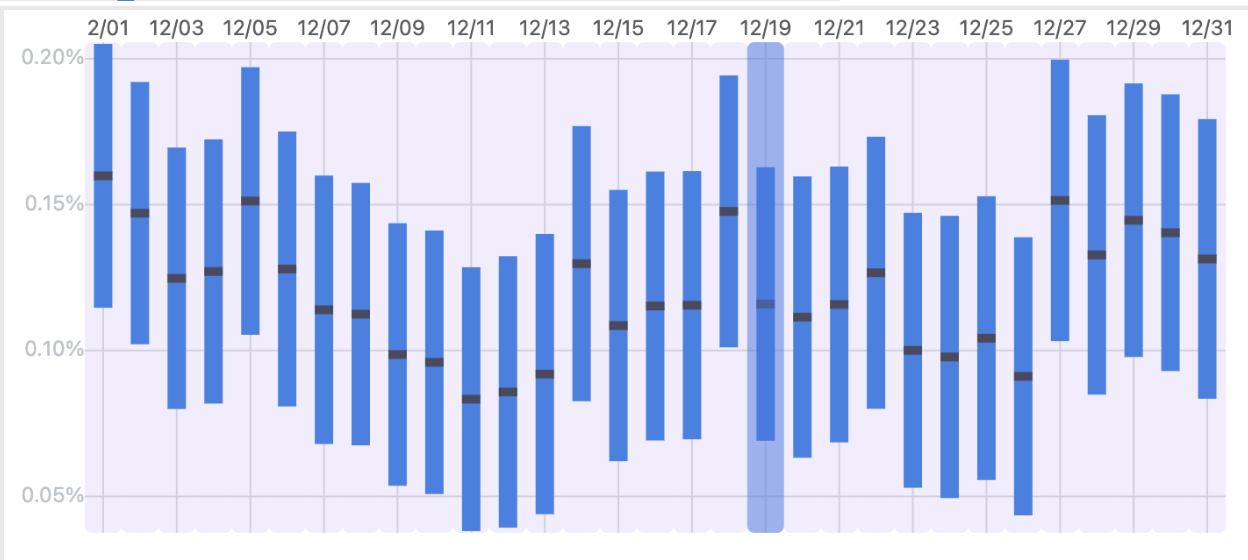}
     \caption{The backtest results indicate that the metrics improvement achieved by the selected policy remains consistent over a period of one month.}
    \label{fig:backtest}
\end{figure}

As illustrated in \autoref{fig:stability}, to validate the efficacy of our stability governance, we conducted a controlled selection from a randomized experiment with four policy candidates. After filtering out high-variance candidates, the remaining policy (\textit{Best for Metric 2}) demonstrated consistent metric improvement over a one-month period~\autoref{fig:backtest}.


\subsection{Broad Adoption and Real-World Impact}

\begin{table}[ht]
  \caption{Online performance improvement of GEARS in different products at Meta. A dash (—) indicates that the specific metric was not applicable or not the primary optimization target for that surface.}
  \label{tab:major_launches}
  \centering
  \small
  \resizebox{.49\textwidth}{!}{
      \begin{tabular}{lccc}
        \toprule
        \textbf{Domain} &
        {\textbf{Metric 1 (\%)}} &
        {\textbf{Metric 2 (\%)}} &
        {\textbf{Metric 3 (\%)}} \\
        \midrule
        Surface 1                    & 0.14   & \multicolumn{1}{c}{—} & \multicolumn{1}{c}{—} \\
        Surface 2                    & \multicolumn{1}{c}{—} & \multicolumn{1}{c}{—} & 0.08 \\
        Surface 3                         & 0.10   & 0.089  & \multicolumn{1}{c}{—} \\
        Surface 4   & 0.10   & \multicolumn{1}{c}{—} & \multicolumn{1}{c}{—} \\
        Surface 5              & 0.042  & \multicolumn{1}{c}{—} & \multicolumn{1}{c}{—} \\
        Surface 6           & 0.011  & 0.017  & 0.08 \\
        Surface 7                    & 0.0406 & \multicolumn{1}{c}{—} & \multicolumn{1}{c}{—} \\
        Surface 8                         & 0.13   & 0.37   & \multicolumn{1}{c}{—} \\
        Surface 9                         & \multicolumn{1}{c}{—} & 0.044  & 0.02 \\
        \bottomrule
      \end{tabular}
    }
\end{table}

~\autoref{tab:major_launches} demonstrates that GEARS delivers improvements across various products. In each deployment, we observe clear and measurable gains in the key metrics.
Taken together, these results support the paper’s claim that GEARS is an effective, general-purpose mechanism for turning cohort-level personalization into repeatable efficient general Agent framework, rather than a one-off optimization tied to a single surface or a single metric.

\subsection{Cost Analysis}
GEARS operates entirely offline and does not affect user-facing
latency. A typical session consumes approximately 3.9K input tokens
and 35.8K output tokens, costing roughly \$3.75 per run. At a scale
of 400 experiments per year, the annual LLM cost is approximately
\$1,500. Prior to GEARS, the equivalent manual workflow (launching
HTE flows, querying feature definitions, evaluating feasibility)
required 1--2 weeks of engineering time; GEARS completes this
pipeline in 1--2 days.

\section{Practical Evaluation}
\label{sec:analysis}
We present an anonymized case study to illustrate how GEARS can recommend actionable optimization policies under multi-objective constraints with minimal manual iteration. In \S~\ref{sec:video_case_study}, we study a large-scale recommendation setting where two primary objectives exhibit a consistent trade-off, and show how GEARS discovers cohorts that admit differentiated treatments while respecting additional guardrail metrics.

\subsection{Complex Trade-off Optimization in Large-Scale Recommendation}
\label{sec:video_case_study}

\begin{table}[htbp]
\centering
\caption{Quantitative results for Baseline Treatment Arms. The table reports the percentage lift relative to the control group ($\text{mean} \pm \text{standard error}$).} 
\label{tab:fb_video_baseline}
\small
\resizebox{.49\textwidth}{!}{
    \begin{tabular}{lcc}
    \toprule
    \textbf{Treatment Arm} & \textbf{Metrics 1} & \textbf{Metrics 2} \\ 
    \midrule
    \begin{tabular}[c]{@{}l@{}} Treatment 1\end{tabular} & 
    \textcolor{red!80!black}{$-0.049\% \pm 0.043$} \ding{55} & 
    \textbf{\textcolor{green!60!black}{+0.282\% $\pm$ 0.074}} \ding{51} \\ 
    \addlinespace[10pt]
    \begin{tabular}[c]{@{}l@{}} Treatment 2\end{tabular} & 
    \textbf{\textcolor{green!60!black}{+0.036\% $\pm$ 0.034}} \ding{51} & 
    \textcolor{red!80!black}{$-0.289\% \pm 0.073$} \ding{55} \\ 
    \bottomrule
    \end{tabular}
}
\end{table} 

In large-scale recommendation systems, improving one engagement objective often comes at the expense of another, creating a persistent multi-objective optimization challenge where globally uniform treatments can lead to near zero-sum outcomes.

As shown in \autoref{tab:fb_video_baseline}, two competing global variants highlight this tension: a treatment improves Metric 1 but degrades Metric 2, while another treatment improves Metric~2 at the cost of Metric~1.

GEARS addresses this by replacing manual slice-and-dice analyses with an agentic personalization workflow. Given a high-level natural language prompt (e.g., \textit{How can I find the tradeoff between metric 1 and metric 2 for this experiment}), the agent iteratively explores a high-dimensional cohort space, proposes candidate segments, and validates them against both primary objectives and guardrail metrics. The outcome is a set of cohort-specific policies where at least one primary objective improves with statistical significance while the other objectives remain neutral within acceptable bounds. ~\autoref{tab:case_topk} reports the top-3 policies recommended by
GEARS. All three achieve statistically significant gains on Metric~2
while maintaining neutrality on Metric~1.

GEARS has found that different types of users react differently to changes in the mix of content they see. For example, users who are very active benefit more from the first treatment, which improves one key metric but doesn’t affect another much. On the other hand, less active users prefer the second treatment, which helps them become more engaged.

\begin{table}[htbp]
  \centering
  \caption{Top-3 policies recommended by GEARS for the case study.
  All three achieve statistically significant gains on Metric~2
  while maintaining neutrality on Metric~1
  (95\% CI includes zero).}
  \label{tab:case_topk}
  \small
  \resizebox{.48\textwidth}{!}{
  \begin{tabular}{c cc cc}
  \toprule
  & \multicolumn{2}{c}{\textbf{Metric 1}}
  & \multicolumn{2}{c}{\textbf{Metric 2}} \\
  \cmidrule(lr){2-3} \cmidrule(lr){4-5}
  \textbf{Policy} & Lift (\%) & 95\% CI & Lift (\%) & 95\% CI \\
  \midrule
  1 & $-0.032$ & $\pm 0.057$ & $\mathbf{+0.350}$ & $\pm 0.092$ \\
  2 & $-0.029$ & $\pm 0.057$ & $\mathbf{+0.349}$ & $\pm 0.092$ \\
  3 & $-0.025$ & $\pm 0.057$ & $\mathbf{+0.338}$ & $\pm 0.092$ \\
  \bottomrule
  \end{tabular}
  }
  \end{table}

Deploying the resulting cohort-targeted policy yields a statistically significant lift on the prioritized metric while maintaining neutrality on the competing metric. Operationally, GEARS automated what was previously a multi-week, expert-driven discovery process, and leverages the multi-agent system to explain how the decision was made and the rationale behind the decision in current ranking context. This significantly improves system efficiency and enabling broader exploration of the policy space through effective human-AI collaboration.

\section{Conclusion}

We introduced GEARS, an orchestration framework that decomposes
ranking policy optimization into programmatic candidate generation, domain-grounded interpretation via specialized agent skills, and deterministic lifecycle governance. Evaluation across 100 structured instructions and nine production surfaces shows that each component contributes where it is most effective: code execution drives performance on structured selection tasks, specialized skills provide gains on tasks requiring domain knowledge, and governance hooks filter out policies relying on unstable features. Deployment reduces the engineering cycle from weeks to days at an annual LLM cost under \$1{,}500.

\begin{acks}
We gratefully acknowledge Tianyao Sun and Zhengyu Chen for their valuable contributions and support.
\end{acks}

\bibliographystyle{ACM-Reference-Format}
\bibliography{reference}


\end{document}